\documentclass[letterpaper]{article} 
\usepackage[]{aaai2026}  
\usepackage{times}  
\usepackage{helvet}  
\usepackage{courier}  
\usepackage[hyphens]{url}  
\usepackage{graphicx} 
\usepackage{natbib}  
\usepackage{caption} 
\usepackage{algorithm}
\usepackage{algorithmic}
\usepackage{booktabs} 
\usepackage{pifont}
\usepackage{newfloat}
\usepackage{listings}
\DeclareCaptionStyle{ruled}{labelfont=normalfont,labelsep=colon,strut=off} 
\floatstyle{ruled}
\newfloat{listing}{tb}{lst}{}
\floatname{listing}{Listing}
\title{3D4D: An Interactive, Editable, 4D World Model via 3D Video Generation}

\author {
    Yunhong He\textsuperscript{*,†},
    Zhengqing Yuan\textsuperscript{\rm 2,*},
    Zhengzhong Tu\textsuperscript{\rm 3},
    Yanfang Ye\textsuperscript{\rm 2},
    Lichao Sun\textsuperscript{\rm 1}
}
\affiliations {
    \vspace{10pt}
    \textsuperscript{\rm 1}\textbf{Lehigh University}, \textsuperscript{\rm 2}\textbf{University of Notre Dame}, \textsuperscript{\rm 3}\textbf{Texas A\&M University}\\
    \vspace{-20pt}
}

\begin{document}

\maketitle

\renewcommand{\thefootnote}{\fnsymbol{footnote}}
\footnotetext[1]{Equal contribution}
\footnotetext[2]{Yunhong He is an independent undergraduate student, remotely working with Lichao Sun.}

\begin{abstract}
We introduce 3D4D, an interactive 4D visualization framework that integrates WebGL with Supersplat rendering. It transforms static images and text into coherent 4D scenes through four core modules and employs a foveated rendering strategy for efficient, real-time multi-modal interaction. This framework enables adaptive, user-driven exploration of complex 4D environments. The project page and code are available at \url{https://yunhonghe1021.github.io/NOVA/}.
\end{abstract}


\section{Introduction}

Advances in Generative Model and Multi-modal Learning have enabled more intuitive text-driven and multi-modal interactive generation, enhancing the usability and performance of 4D content generation \cite{fei-etal-2024-xnlp,liu2023visual,rombach2022high,sun2024bora,yuan2024mora,wei2025unsupervised,li2023metaagents,yan2024biomedicalsam2segment,yuan2024vit,singer2023text}. Yet, conventional WebGL-based frameworks remain limited in their ability to handle real-time 4D rendering and fine-grained temporal navigation, often struggling with high computational costs, latency, and scalability issues \cite{xu20244k4d, li2024user, abbas2025realtime, jang2022scene}. More critically, existing systems fall short of providing genuinely interactive 4D environments, as technologies that seamlessly combine high-performance rendering with user interactivity are still largely absent~\cite{miao2025advances,zhou2025libero}.

\begin{figure}[t]
    \centering
    \includegraphics[width=\linewidth]{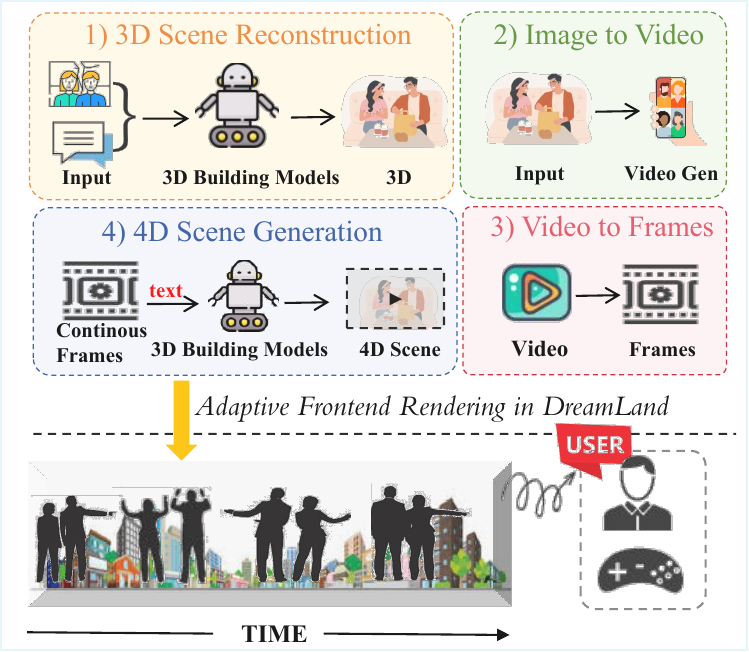}
    \vspace{-20pt}
·    \caption{Overview of 3D4D pipeline. The system integrates multi-modal inputs with real-time 4D rendering to support interactive exploration.}
    \label{fig:1}
    \vspace{-15pt}
\end{figure}

To address these challenges, we introduce 3D4D, an interactive 4D visualization framework that integrates WebGL with Supersplat~\cite{playcanvas2025supersplat_editor} rendering to support real-time interaction and editing. As illustrated in Figure~\ref{fig:1}, the backend comprises four core modules—3D Scene Reconstruction, Image-to-Video Synthesis, Video-to-Frame Decomposition, and 4D Scene Generation—that transform static images and text into temporally coherent 4D scenes for adaptive, user-driven exploration. This paper primarily focuses on the frontend component, which enables real-time interaction, editing, and visualization within the generated 4D environment.

\begin{figure}[th] 
    
    \centering
    \includegraphics[width=1\linewidth]{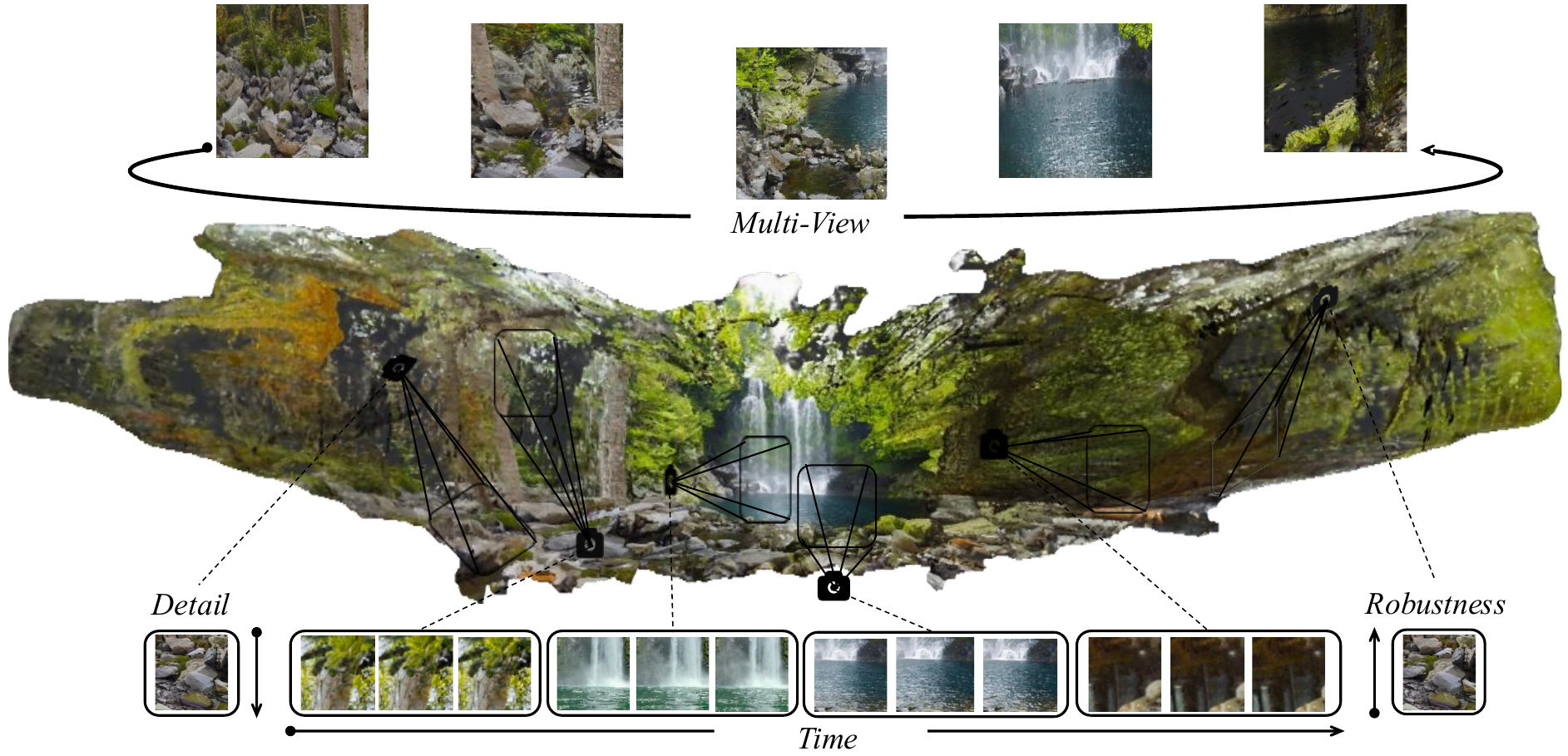}
    \vspace{-10pt}
    \caption{
Illustrative input–prompt pair and evaluation axes. upper left: the single panoramic photograph fed to DreamGen. lower left: its accompanying natural-language prompt requesting. This pair serves as a running example for visualization results, where the generated 4D scene is assessed on the three WorldScore axes—\emph{Controllability}, \emph{Quality}, and \emph{Dynamics}.}
    \label{fig:2}
    \vspace{-15pt}
\end{figure}

3D4D further introduces a foveated rendering strategy inspired by human peripheral vision, progressively refining semantically important regions identified by a vision-language model while approximating low-saliency areas. This design aims to maintain semantic alignment and visual consistency while reducing GPU memory usage and latency, thereby enabling real-time 4D interaction even under constrained computational resources. An illustrative demo of 3D4D’s interactive 4D scene generation is shown in Figure~\ref{fig:2}. For a more detailed illustration, please refer to our demonstration.

\section{System Framework}
Our front-end is developed based on WebGL and Supersplat rendering technologies to achieve high-performance visualization of 4D environments. Since the standard WebGL framework does not support fine-grained temporal interaction, we have developed several customized functions on this basis, enabling users to manipulate, edit, and analyze the generated 4D scenes in real time.

\subsection{Design of Frontend}


The frontend, developed with WebGL and Supersplat rendering, provides high-performance visualization and real-time editing of 4D environments. When a user inputs an image and a text prompt, the system triggers the backend to reconstruct the corresponding 3D scene and synthesize temporally coherent frames. These outputs are converted into multiple PLY point clouds and streamed to the frontend, where they are rendered sequentially or looped to form a continuous 4D video~\cite{deng2024vg4d,casas20144d,wu2024sc4d}.

The interface allows users to dynamically explore the generated scene by adjusting camera pose, playback speed, and frame rate through an interactive timeline. Fine-grained temporal navigation enables users to balance visual quality and efficiency during analysis. Furthermore, the frontend supports direct scene editing through tools such as rectangle, brush, polygon, lasso, and sphere selection, enabling precise manipulation of objects and regions. All user interactions are synchronized with the backend via an API, ensuring seamless integration of rendering, computation, and user control.
\begin{figure}[t]
\includegraphics[width=\linewidth]{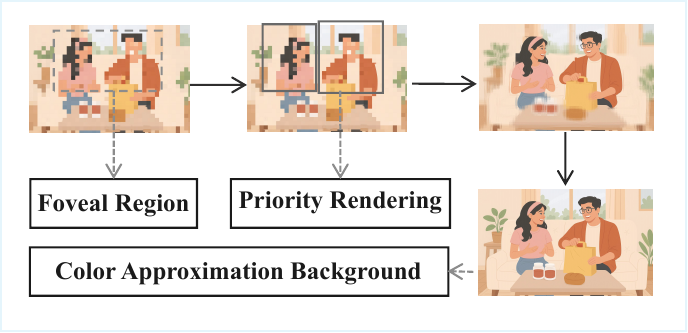}
\vspace{-25pt}
\caption{3D4D’s adaptive frontend rendering: semantically important regions are rendered in high resolution, while peripheral areas are approximated to reduce cost.}
\label{fig:3}
\vspace{-15pt}
\end{figure}

\subsection{Rendering Video}
When the user clicks ``Rendering Video'', the system initiates a VLM-guided rendering pipeline that transforms uploaded PLY scenes into high-quality videos. After loading one or multiple Gaussian Splat point clouds and interpolating camera trajectories from user-defined keyframes, a Vision-Language Model (VLM)~\cite{yuan2023tinygpt} like Qwen2.5vl~\cite{Qwen2VL} analyzes each frame to generate an importance map identifying semantically critical regions such as humans or moving objects.  As illustrated in Figure~\ref{fig:3}During rendering, the WebGL shader adaptively allocates resources, rendering foveal regions at full precision while applying blurred, low-cost shading to background areas, thereby reducing GPU load without compromising perceptual quality. Each frame is captured from the framebuffer, temporally smoothed, and encoded in real time through the browser’s MediaRecorder into \texttt{.webm} or \texttt{.mp4} format. This fully client-side pipeline achieves semantic-aware, real-time 4D video generation that balances visual fidelity and computational efficiency.

\subsection{Evaluation}
\noindent\textbf{Metrics.}~~We evaluate our system using both performance and efficiency metrics. Performance metrics include CLIP Score (CS), calculated as the CLIP \cite{radford2021learning} similarity between the textual scene prompt and the rendered image, indicating the semantic alignment quality; and CLIP Consistency (CC), measured by the cosine similarity between the CLIP embeddings of each novel view image and the central reference view, evaluating visual consistency across different viewpoints. Efficiency metrics include frames per second (fps), measuring the rendering speed, and a binary indicator of real-time interaction system.

\begin{table}[t]
\centering
\small
\begin{tabular}{lcc}
\toprule
\textbf{Model} & \textbf{CC} & \textbf{CS} \\
\midrule
WonderJourney~\cite{yu2024wonderjourney} & 27.34 & 0.9544 \\
LucidDreamer & 26.72 & 0.8972 \\
Text2Room~\cite{hollein2023text2room} & 24.50 & 0.9035 \\
WonderWorld & 29.47 & 0.9948 \\
SV4D~\cite{xie2024sv4d} & 30.29 & 0.8856 \\
4D-fy~\cite{bahmani20244d} & 11.23 & 0.6147 \\
\textbf{3D4D (Ours)} & \textbf{30.40} & \textbf{0.9951} \\
\bottomrule
\end{tabular}
\caption{Evaluation results for CLIP Consistency (CC) and CLIP Score (CS) across different models.}
\vspace{-15pt}
\label{tab:evaluation_results}
\end{table}

\begin{table}[h]
\small
\centering
\begin{tabular}{lcc}
\toprule
\textbf{Model} & \textbf{FPS} & \textbf{Real-time Interaction} \\
\midrule
SVD-4D \cite{xie2024sv4d} & 40 & \ding{55} \\
4D-fy \cite{bahmani20244d} & 16 & \ding{55} \\
\textbf{3D4D (Ours)} & \textbf{60} & \ding{51} \\
\bottomrule
\end{tabular}
\caption{Performance comparison in terms of frame rate (FPS) and real-time interaction capability.}
\label{tab:performance_metrics}
\vspace{-10pt}
\end{table}

\noindent\textbf{Results.}~~Our proposed system, 3D4D, outperforms existing methods across both performance and efficiency metrics. Specifically, as demonstrated in Table~\ref{tab:evaluation_results}, 3D4D achieves the highest CC of 30.40 and the best CS of 0.9951, indicating superior semantic alignment and visual consistency compared to other models. Moreover, Table~\ref{tab:performance_metrics} highlights 3D4D's substantial improvement in rendering speed, delivering 60 frames per second (fps), and uniquely supporting real-time interaction, a critical difference for other methods.

\bibliography{aaai2026}

\end{document}